\documentclass[conference]{IEEEtran}
\IEEEoverridecommandlockouts
\usepackage{cite}
\usepackage{url}
\usepackage{amsmath,amssymb,amsfonts}
\usepackage{amsthm}
\theoremstyle{definition}
\newtheorem{definition}{Definition}
\theoremstyle{remark}
\newtheorem*{example}{Example}
\usepackage{algorithmic}
\usepackage{graphicx}
\usepackage{textcomp}
\usepackage{xcolor}
\def\BibTeX{{\rm B\kern-.05em{\sc i\kern-.025em b}\kern-.08em
    T\kern-.1667em\lower.7ex\hbox{E}\kern-.125emX}}
    
\usepackage{tikz}
\usepackage{tabularx}
\usepackage{array}
\newcolumntype{M}[1]{>{\centering}m{#1}}

\usepackage{prettyref}
\newrefformat{tab}{Table~\ref{#1}}
\newrefformat{def}{Def.~\ref{#1}}
\newrefformat{fig}{Fig.~\ref{#1}}
\newrefformat{lem}{Lemma~\ref{#1}}
\newrefformat{sec}{Sect.~\ref{#1}}
\newrefformat{ssec}{Sect.~\ref{#1}}
\newrefformat{line}{l.\ref{#1}}
\def\pref{\prettyref}

\usepackage{listings}
\lstdefinelanguage{ASP}{%
literate={:-}{$\leftarrow$}1,
morekeywords={not},
numbers=left,
escapeinside={\%}{\^^M}
}

\renewcommand{\thelstnumber}{\the\value{lstnumber}}
\def\B{{\mathbb B}}
\def\T{(\B\cup\{*\})}
\def\range#1#2{\{#1, \ldots, #2\}}
\def\DEF{:=}

\def\step{\rightarrow}
\def\astep{\rightarrow_a}
\def\sstep{\rightarrow_s}
\def\reach{\step^*}

\begin{document}

\title{Synthesis of Boolean Networks from Biological Dynamical Constraints
    using Answer-Set Programming
\thanks{The authors ackowledge the support from ITMO Cancer and from
    the French Agence Nationale pourla Recherche (ANR), in the context of ANR-FNR project
“AlgoReCell” ANR-16-CE12-0034}}

\author{\IEEEauthorblockN{St\'ephanie Chevalier}
\IEEEauthorblockA{LRI, CNRS, U. Paris-Sud\\
U. Paris-Saclay, France\\
stephanie.chevalier@lri.fr}
\and
\IEEEauthorblockN{Christine Froidevaux}
\IEEEauthorblockA{LRI, CNRS, U. Paris-Sud\\
U. Paris-Saclay, France\\
christine.froidevaux@lri.fr}
\and
\IEEEauthorblockN{Lo\"ic Paulev\'e}
\IEEEauthorblockA{LaBRI, CNRS, U. Bordeaux\\
Bordeaux INP, France\\
loic.pauleve@labri.fr}
\and
\IEEEauthorblockN{Andrei Zinovyev}
\IEEEauthorblockA{Institut Curie, INSERM\\
U. PSL, Mines ParisTech, France\\
andrei.zinovyev@curie.fr}
}

\maketitle

\begin{abstract}
Boolean networks model finite discrete dynamical systems with complex behaviours.
The state of each component is determined by a Boolean function of the state of (a subset of) the
components of the network.

This paper addresses the synthesis of these Boolean functions from constraints on their domain and
emerging dynamical properties of the resulting network.
The dynamical properties relate to the existence and absence of trajectories between partially
observed configurations, and to the stable behaviours (fixpoints and cyclic attractors).
The synthesis is expressed as a Boolean satisfiability problem relying on Answer-Set Programming with
a parametrized complexity, and leads to a complete non-redundant characterization of the set of solutions.

Considered constraints are particularly suited to address the synthesis of models of cellular
differentiation processes, as illustrated on a case study.
The scalability of the approach is demonstrated on random networks with scale-free structures up to
100 to 1,000 nodes depending on the type of constraints.
\end{abstract}


\section{Introduction}
\label{sec:intro}
The modelling of complex dynamical systems usually requires extensive knowledge on their functioning
to be able to reproduce their observed behaviours.
For most physical and biological systems, such a knowledge is out of reach.
In systems biology, the vast majority of (if not all) models involved trial error approaches with
arbitrary choices for specifying the rules of the model, until its dynamics fits with the desired
behaviour.

The synthesis of dynamical models aims at providing an automatic way of designing models that
satisfy constraints derived from knowledge on the structure and on the behaviour of the system,
and potentially gives insight into the diversity of such models.

In this paper, we address the synthesis of Boolean Networks (BNs) from dynamical properties derived from partial and
discrete-time observations of the system.
BNs model the dynamics of a finite set of nodes having binary states.
The possible evolution of these configurations are computed according to a collection of Boolean
functions and an update semantics.
BNs are close to 1-bounded Petri nets \cite{concurrency-in-BNs}, and are extensively applied to
model the complex dynamics of biological networks.
We consider positive and negative reachability properties, i.e., the ability (or impossibility) for
the model to evolve from one configuration to another;
and long-run properties, i.e., on configurations that are eventually reached after an infinite
amount of time.
The domain of Boolean functions composing the candidate BNs is typically delimited by a given
influence graph (often called \emph{Prior Knowledge Network}), which specifies for each node the
variables that can be used in its Boolean function.

These properties are motived by the modelling of cellular differentiation processes.
Starting from a multi-potent (stem) state, cells progressively specialize into specific types.
Various biological experimentation techniques measure the activities of certain genes during
the differentiation processes (at different times).
From these observations can then be derived positive reachability properties to reproduce the
sequence of observed states; but also attractor properties when observations have been performed
in stabilized cells.
Finally, negative reachability properties enable to model bifurcations inherent in the
differentiation process: once a cell enters a particular branch of differentiation, it is impossible
for it to reach cell types related to the other branches.

In the literature, the synthesis of BNs subject to static and dynamical properties derived from
partial and discrete-time observations essentially
splits into either evolutionary optimization algorithms, or satisfiablility problems.
Methods of the former category, such as \cite{Terfve2012,Dorier2016}, couple genetic algorithms to
explore the model space together with simulations to assess positive reachability and attractor
properties.
In practice, they allow addressing networks between 20-40 nodes.
Such approaches do not guarantee terminating, nor finding a globally optimal model.
Moreover, they offer a very limited access to the space of solutions of the synthesis problem.
On the other hand, \cite{Caspots-BioSystems16} uses Answer-Set Programming, and \cite{yordanov2016a}
Satisfiability Modulo Theory (SMT), to express the synthesis problem.
Such approaches enable the exhaustive enumeration of all the solutions, potentially subject to optimization
criteria.
In \cite{Caspots-BioSystems16}, only positive reachability properties are considered using
model-checking, and have been applied to network up to 80 nodes.
In \cite{yordanov2016a},
both positive reachability and fixpoint properties are supported, but only
a particular subset of candidate Boolean functions are explored.
Applications show scalability up to 20-40 nodes, with the synchronous semantics.

In this paper, we consider the logical synthesis of BNs from attractors, positive, and \emph{negative}
reachability properties using Answer-Set Programming (ASP), giving a complete characterization of
the solutions.
The considered dynamical constraints can be typically derived from the observation of cellular
differentiation processes.
We rely on the most permissive semantics of BNs, which offers both a correct abstraction of
non-Boolean systems (as for biological systems), and a high scalability for the verification of
dynamical properties.

\section{Background}
\label{sec:bg}

In this section, we formally define BNs, their influence (causal) graphs, and dynamical
properties related to stability (trap spaces, attractors) and trajectories (reachability).
Finally, we give a short introduction to Answer-Set Programming.

\subsection{Boolean Networks}
\label{sec:bg-BNs}

A \emph{Boolean network} (BN) of dimension $n$ is a function
\begin{equation}
f:\B^n\to\B^n
\end{equation}
where $\B\DEF\{0,1\}$.
For all $i\in\range 1n$, $f_i:\B^n\to\B$ denotes the \emph{local function} of the $i$-th component.
A vector $x\in\B^n$ is called a \emph{configuration} of the BN $f$.
The set of components which differ between two configurations $x,y\in\B^n$ is denoted by
$\Delta(x,y)\DEF\{ i\in\range1n\mid x_i\neq y_i\}$.

A BN $f$ is said \emph{locally monotonic} whenever each of its local functions is monotonic
(this does not imply $f$ monotonicity).
Intuitively, when expressing the local functions using propositional logic, local monotonicity
imposes that a variable appears always with the same sign in a minimal normal form.

\begin{figure}[!t]
\begin{minipage}{0.5\linewidth}
    \centering
\begin{align*}
f_1(x) &\DEF \neg x_2\\
f_2(x) &\DEF\neg x_1\\
f_3(x) &\DEF \neg x_1 \wedge x_2
\end{align*}
\end{minipage}
\hfill
\begin{minipage}{0.4\linewidth}
\centering
\begin{tikzpicture}[node distance=1.5cm,font=\footnotesize]
\node (1) {1};
\node[below right of=1] (3) {3};
\node[below of=1] (2) {2};
\path[->]
(1) edge[bend left,-|] (2)
(2) edge[bend left,-|] (1)
(1) edge[-|] (3)
(2) edge (3)
;

\end{tikzpicture}
\end{minipage}

\caption{Example of Boolean network \(f\) and its influence graph \(G(f)\) where
positive edges are with normal tip and
negative edges are with bar tip.}
\label{fig:bn-example}
\end{figure}
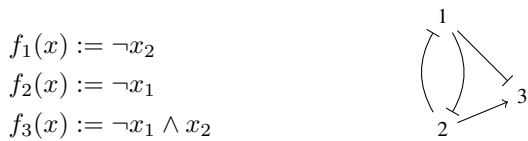

\pref{fig:bn-example} is an example of locally-monotonic BN with $n=3$.

\subsection{Influence Graph}
\label{sec:bg-IG}

For each component $i\in\range 1n$, $f_i$ typically depends only on a subset of components of the
BN.
The \emph{influence graph} (also called interaction or causal graph)
summarizes these dependencies by having an edge from node $j$ to $i$ if $f_i$
depends on the value of $j$.
Formally, $f_i$ depends on $j$ if there exists a configuration $x\in \B^n$ such that
$f_i(x)$ is different from $f_i(x')$ where $x'$ is $x$ having solely the
component $j$ different ($x'_j = \lnot x_j$).
Moreover, assuming $x_j=0$ (therefore $x'_j=1$), we say that $j$ has a positive
influence on $i$ (in configuration $x$) if $f_i(x) < f_i(x')$, and a negative influence
if $f_i(x) > f_i(x')$.
It is possible that a node has different signs of influence on $i$ in different
configurations (leading to non-monotonic $f_i$).
Remark that different BNs can have the same influence graph.

\begin{definition}
Given a BN $f$ of dimension $n$, its \emph{influence graph} $G(f)$ is a directed graph $(\range
1n,E_+,E_-)$ with \emph{positive} and \emph{negative} edges such that
\((j,i)\in E_+\) (resp. \((j,i)\in E_-\)) iff 
\(\exists x,y\in \B^n\) s.t. \(\Delta(x,y)=\{j\}\), \(x_j<y_j\), 
and \(f_i(x) < f_i(y)\) (resp. \(f_i(x) > f_i(y)\)).
\end{definition}

Given two influence graphs $\mathcal G=(\range 1n,E_+,E_-)$ and $\mathcal G'=(\range 1n,E'_+,E'_-)$,
we say that \(\mathcal G\) is a subgraph of $\mathcal G'$, denoted by $\mathcal G\subseteq\mathcal G'$
iff $E_+\subseteq E'_+$ and \(E_-\subseteq E'_-\).

\pref{fig:bn-example} (right) shows the influence graph of the BN example.

\subsection{Meta-configurations and trap spaces}

The results presented in this paper extensively use the notion of
meta-configurations, which denote hypercubes within \(\B^n\), i.e., a set of components being
fixed to a Boolean state, and the others being free (noted with \(*\)).
\begin{definition}
    A \emph{meta-configuration} \(h\) of dimension \(n\) is a vector in $\T^n$.
    The set of its associated configurations is denoted by
    \(c(h)\DEF \{x\in\B^n\mid \forall i\in\range 1n, h_i\neq *\Rightarrow x_i=h_i\}\).

    Given two meta-configurations \(h,h'\in\T^n\), \(h\) is \emph{smaller} than \(h'\)
    iff
    \(\forall i\in\range1n, h'_i\neq*\Rightarrow h_i=h'_i\).
\end{definition}

Trap spaces of a BN \(f\) are special cases of meta-configurations which are closed by \(f\):
\begin{definition}\label{def:trap-space}
    A \emph{trap space} of a BN \(f\) of dimension \(n\) is
    a meta-configuration \(t\in\T^n\)
    such that
    \(\forall x \in c(t), f(x)\in c(t)\).
\end{definition}
\noindent
A trap space is \emph{minimal} if there is no smaller trap space.
Remark that if \(x\in\B^n\) is a fixpoint of \(f\), i.e., \(f(x)=x\), then \(x\) is a (minimal) trap space
(hypercube of dimension \(0\)).

Finally, given a BN \(f\) of dimension \(n\) and a set of components \(L\subseteq\range1n\), a
\(L\)-constrained trap space is defined similarly, except that the closure is ensured only for the
components \emph{not} in \(L\):
\begin{definition}\label{def:constrained-trap-space}
    A \emph{\(L\)-constrained trap space} of a BN \(f\) of dimension \(n\) with
    \(L\subseteq\range1n\)
    is a meta-configuration \(w\in\T^n\)
    such that
    \(\forall i\in \range 1n\setminus L\),
    either \(w_i=*\), or
    \(\forall x\in c(w)\),
    \(w_i=f_i(x)\).
\end{definition}
\noindent
Remark that \(\emptyset\)-constrained trap spaces are equivalent to trap spaces.

In the following, we will often rely on \emph{smallest} (constrained) trap spaces
\emph{containing} a configuration \(x\).
These smallest meta-configurations can be obtained by transfinite iterations of functions
\(\T^n\to \T^n\) enlarging meta-configurations to satisfy the trap conditions, initially applied to \(x\).
For instance, the smallest \(L\)-constrained trap space containing \(x\) can be obtained by the
transfinite iteration of \(e\) initially applied to \(x\),
where \(e(h)=h'\) verifies \(\forall i\in\range1n\),
\(h'_i = *\) if \(i\notin L\) and \(\exists x\in c(h): f_i(x)\neq x_i\), otherwise
    \(h'_i=h_i\).

\begin{example}
    The meta-configuration \(01*\) is a trap space of the BN \(f\) of \pref{fig:bn-example};
    \(c(01*)=\{010,011\}\).
    The meta-configuration \(1\!*\!0\) is a \(\{1\}\)-constrained trap space of \(f\),
    it is the smallest \(\{1\}\)-constrained trap space containing \(110\),
    and it is not a trap space, nor the smallest \(\{1\}\)-constrained containing \(100\).
\end{example}

\subsection{Reachability}
\label{sec:bg-reachability}

Given two configurations $x,y\in\B^n$, $y$ is \emph{reachable} from $x$, noted $x\reach y$, if there
exists a possible evolution of the configuration $x$, according to the BN $f$, which leads to $y$.

Numerous semantics of BNs have been defined in the literature \cite{Kauffman69,Thomas73,Aracena09},
the most prominent being the \emph{synchronous update mode}, where $\reach$ is the transitive
closure of the binary relation $\sstep\,\subseteq\B^n\times\B^n$ with $x\sstep y$ iff $f(x)=y$,
i.e., all components get updated simultaneously in one step;
and the \emph{asynchronous update mode}, where $\reach$ is the transitive closure of the
binary relation 
$\astep\,\subseteq\B^n\times\B^n$ with $x\astep y$ iff $\forall i\in\Delta(x,y), y_i=f_i(x)$,
i.e., any number of components gets updated (non-deterministically) in one step.

However, all the update modes of BNs are inconsistent abstractions of non-Boolean systems dynamics
\cite{beyond-general}, i.e., they both introduce spurious reachability properties and miss
reachability properties actually verified in more concrete quantitative specifications.
This constitutes a prime issue for BN synthesis as it may lead to reject valid models.

The \emph{most permissive} semantics of BNs has been recently introduced to address this issue
\cite{concurrency-in-BNs,mpbn-tr}.
This semantics is currently the only one known which guarantees that its reachability properties are a
correct over-approximation of reachability properties in any quantitative refinement of
the BN, with any update mode.

In this paper, we focus on most permissive BNs.
The reachability property \(x\reach y\) can then be characterized with the smallest
constrained trap spaces containing \(x\):
\(y\) has to be contained in one of such meta-configurations \(w\), and in the case a component \(i\) is free 
(\(w_i=*\)) whereas \(x_i=y_i\), then there should exist a configuration \(z\in c(w)\) such that \(f_i(z)=y_i\).
The most permissive reachability is formally defined as follows.
\begin{definition}\label{def:reachability}
    Given a BN \(f\) of dimension \(n\) and two configurations
    \(x,y\in\B^n\),
    \(x\reach y\) if and only if
    there exists
    \(L\subseteq \range 1n\)
    such that
    the smallest \(L\)-constrained trap space \(w\) containing \(x\)
    verifies
    (1) \(y\in c(w)\), and
    (2) \(\forall i\in \range1n\setminus L\) where \(x_i=y_i\) and \(w_i=*\),
    \(\exists z\in c(w)\) s.t \(f_i(z)=y_i\).
\end{definition}

Deciding \(x\reach y\) in locally-monotonic BNs of dimension \(n\) is in PTIME --
NP-complete for general BNs -- instead of PSPACE-complete with classical update modes%
\cite{concurrency-in-BNs,mpbn-tr}.

\begin{example}
In the BN \(f\) of \pref{fig:bn-example},
    \(000\reach 111\),
    \(110\reach 000 \reach 110\) (\(L=\emptyset\)),
    but \(010\not\reach 100\) (\(w=01*\) with \(L=\emptyset\)).
	In the BN \(g:\B^3\to\B^3\) with
    \(g_1(x)\DEF 1\),
    \(g_2(x)\DEF x_1\wedge x_3\) and \(g_3(x)\DEF\neg x_2\),
    \(011\reach 000\) \((L=\{1\}, w=0**)\),
    but \(001\not\reach 010\) \((\)either \(1\notin L\), then \(\nexists z \in c(w):f_1(z)=0\), or \(1\in L\), then \(w=001)\).
\end{example}

\subsection{Attractors}
\label{sec:bg-attractors}

The long-run behaviour of BNs is characterized by so-called \emph{attractors}, which are the
smallest sets of configurations closed by the reachability relation:
\begin{definition}
    An \emph{attractor} of a BN \(f\) of dimension \(n\)
    is a set of configurations \(A\subseteq\B^n\) such that
    \(\forall x,y\in A, x\reach y\) and \(y\reach x\),
    and \(\forall x\in A, z\in\B^n\),
    \(x\reach z \Rightarrow z\in A\).

    The set of attractors of \(f\) is denoted by \(\mathcal A(f)\).
\end{definition}
We usually distinguish two kinds of attractors: the singleton attractors $\{x\}$
corresponding to the fixpoints of the BN (\(f(x)=x\)); and the cyclic attractors.

With the most permissive semantics, attractors match exactly with the \emph{minimal} trap spaces of
\(f\) \cite{mpbn-tr}.

\begin{example}
    The BN \(f\) of \pref{fig:bn-example} has two attractors, being, in this particular case,
    fixpoints: \(011\) and \(100\).
    The BN \(g\) illustrating \pref{def:reachability} has a single cyclic attractor, being all the
    configurations \(\{100,101,110,111\}\), i.e., the minimal trap space \(1**\).
\end{example}

It is worth noticing that, due to the non-determinism of BN semantics,
one configuration can reach several attractors;
it is the case in the BN \(f\) of \pref{fig:bn-example}, where the configuration \(000\) can reach
the two fixpoints.
This is an important feature of BNs for the modelling of biological differentiation processes.

\subsection{Answer-Set Programming}
\label{sec:bg-ASP}
Answer Set Programming (ASP; \cite{baral02a,gekakasc12a}) is a declarative approach to solving
combinatorial satisfaction problems.
It is close to SAT (propositional satisfiability) \cite{Lin2004} and known to be efficient for
enumerating solutions of NP problems comprising up to tens of millions of variables,
while providing a convenient language for specifying the problem.
We give a very brief overview of ASP syntax and semantics that we use in the next sections; see
\cite{gekakasc12a} for more details.

An ASP program is a Logic Program (LP) being a set
of logical rules with first order logic predicates of the form:
\begin{lstlisting}
$a_0$ :- $a_1$, $\dots$, $a_n$, not $a_{n+1}$, $\dots$, not $a_{n+k}$.
\end{lstlisting}
where $a_i$ are (variable-free) atoms, i.e., elements of the Herbrand base, which is built from all the possible predicates of the LP. The Herbrand base is built by instantiating the LP predicates with 
the LP terms (constants or elements of the Herbrand universe).

Essentially, such a logical rule states that when all $a_1,\dots,a_n$ are true 
and none of $a_{n+1},\dots,a_{n+k}$ can be proven to be true, then $a_0$ has to be true as well.
Whenever $a_0$ is $\bot$ (false), the rule, also called integrity constraint, becomes:
\begin{lstlisting}
:-$a_1$, $\dots$, $a_n$, not $a_{n+1}$, $\dots$, not $a_{n+k}$.
\end{lstlisting}
Such a rule is satisfied only
if the right hand side of the rule is false (at least one of $a_1,\dots,a_n$ is false or at least
one of $a_{n+1},\dots,a_{n+k}$ is true).
On the other hand,
\lstinline|$a_0$ :- $\top$| ($a_0$ is always true) is abbreviated
as
\lstinline|$a_0$|.
A solution (answer set) is a \emph{stable} Herbrand model, that is, a minimal set of true atoms
where all the logical rules are satisfied.

ASP allows using variables (starting with an upper-case) instead of terms/predicates: these
\emph{template} declarations will be expanded to the corresponding propositional logic rules prior to
the solving. For instance, the following ASP program
\begin{lstlisting}
c(X) :- b(X).
b(1).
b(2).
\end{lstlisting}
has as unique solution \lstinline|$\{$b(1)$,$ b(2)$,$ c(1)$,$ c(2)$\}$|.

We also use the notations
\lstinline|a((x;y))| which is expanded to \lstinline|a(x), a(y)|;
\lstinline|#count {X: a(X)}| which is the number of distinct \lstinline|X| for which
\lstinline|a(X)| is true;
\lstinline|$n$ {a(X): b(X)} $m$|
which is satisfied when at least $n$ and at most $m$ \lstinline|a(X)| are true where \lstinline|X| ranges over
the true \lstinline|b(X)|;
and \lstinline|a(X): b(X)| which is satisfied when for each \lstinline|b(X)| true, \lstinline|a(X)| is
true.
If any term follows such a condition, it is separated with \lstinline|;|.
Finally, rules of form
\begin{lstlisting}
{a} :- $\text{\it body}$.
\end{lstlisting}
leave the choice to make \lstinline|a| true whenever the body is satisfied.

\section{Synthesis Problem}
\label{sec:problem}

This paper focuses on the synthesis of BNs from constraints on its influence graph and on its
dynamics, with reachability and attractors properties.

The nature of the constraints is inspired by the modelling of cellular differentiation processes.
In this biological context, a cell population evolves towards various phenotypes, and this behaviour
covers interesting properties both in healthy and pathological context (respectively for studying
embryogenesis and cancer for instance).
Typical experimental data provide partial discrete-time observations of genes and proteins activity
along bifurcating trajectories.
These data can be further statistically processed to provide binary interpretation of the activity of
components at the collected time points and classify them along differentiation branches.
Then, putative components and influences of interest can be extracted from databases and completed
by causal learning from the experimental data.

A (partial) observation \(o\) of a configuration of dimension \(n\) is specified by a set of couples
associating a component to a Boolean value: \(o\subseteq \range1n\times\B\), assuming
there is no \(i\in\range1n\) such that \(\{(i,0),(i,1)\}\subseteq o\).

\def\PR{\mathsf{PR}}
\def\NR{\mathsf{NR}}
\def\FP{\mathsf{FP}}
\def\TP{\mathsf{TP}}

Formally, the synthesis problem we tackle is the following.

\noindent
Given
\begin{itemize}
\item an influence graph \(\mathcal G=\{\range 1n, E_+, E_-)\),
\item \(p\) partial observations $o^1, \ldots, o^p$,
\item sets \(\PR\) and \(\NR\) of couples of indices of observations:
    \(\PR, \NR \subseteq \range1p^2\),
\item subset \(\FP\) of indices of observations:\\
    \(\FP \subseteq \range1p\),
\item a set \(\TP\) associating indices of observations with components:
    \(\TP\subseteq \range1p\times\range1n\),
\end{itemize}
find a BN \(f\) of dimension \(n\) such that
\begin{itemize}
\item \(G(f)\subseteq \mathcal G\),
\item there exist \(p\) configurations \(x^1,\ldots,x^p\) such that:
    \begin{itemize}
        \item(observations) \(\forall m\in\range1p, \forall (i,v)\in o^m, x^m_i=v\),
        \item(positive reachability) \(\forall (m,m')\in\PR, x^m\reach x^{m'}\),
        \item(negative reachability) \(\forall (m,m')\in\NR, x^m\not\reach x^{m'}\),
        \item(fixpoints) \(\forall m\in\FP, f(x^m)=x^m\),
        \item(trap space) \(\forall (m,i)\in\TP, \exists t\in (\B\cup\{*\})^n: \)
            \(t\) is the smallest trap space containing \(x^m\), and \(t_i=x^m_i\).
    \end{itemize}
\end{itemize}

Remark that such a problem can be non-satisfiable depending on the input influence graph and
dynamical properties.
Besides the scalability challenge of such a synthesis problem, desired features include the
\emph{complete} and \emph{non-redundant} characterization of the satisfying BNs.
Completeness is possible as there is a finite number of BNs \(f\) such that \(G(f)\subseteq\mathcal
G\).
Non-redundancy implies that the method should enumerate only among non-equivalent BNs (i.e., where
their values differ for at least one configuration).

\section{Answer-Set Programming Encoding}
\label{sec:encoding}

This section details the ASP encoding of the BN synthesis from constraints on its influence graph
and its dynamics.

The constraints on dynamics relate to the \emph{existence} of configurations which match
their \emph{partial} observations and verify given reachability and stability properties.
A partial observation of configuration \lstinline|X| is specified by \lstinline|obs(X,N,V)|
predicates, where \lstinline|N| and \lstinline|V| denote the component and its observed Boolean value.
Boolean values are encoded as $-1$ for false, and $1$ for true.
The configuration \lstinline|X| is encoded by a set of predicates \lstinline|cfg(X,N,V)|.
If the node \lstinline|N| has been observed, \lstinline|V| is equal to the observed value;
otherwise, its value is chosen:
\begin{lstlisting}[firstnumber=1]
cfg(X,N,V) :- obs(X,N,V).
1 {cfg(X,N,(-1;1))} 1 :- obs(X,_,_), node(N), not obs(X,N,_).
\end{lstlisting}

\subsection{Canonical Domain of Boolean Networks}
\label{sec:encoding-BNs}

The ASP encoding of locally-monotonic BNs compatible with an influence graph faces two difficulties.
First, two different solutions should correspond to two non-equivalent BNs \(f\) and \(f'\), i.e.,
there exists \(x\in\B^n\) such that \(f(x)\neq f'(x)\).
This requires ensuring that solutions match with canonical representations of BNs.
Second, the worst size of the specification of a Boolean function is exponential in the number of
its variables.
Therefore, the encoding should allow specifying a bound on the size of the Boolean function
specification, ideally without bounding the number of variables.

We represent the Boolean functions composing a BN under their Disjunctive Normal Form (DNF), i.e.,
a set of clauses, where clauses are sets of literals, and two distinct clauses have no subset
relation (antichain).
In ASP, we have to encode DNF as lists of clauses, and therefore give an index to each clause.
The canonicity is then ensured by enforcing a total ordering between the clauses.
The maximum number of clauses for a DNF with \(d\) variables is \(\binom d {\lfloor d/2\rfloor}\), 
and our encoding allows specifying a lower number to restrict the set of DNFs to consider, without
limiting the number of variables to consider.

Overall, our encoding of canonical Boolean functions with \(d\) variables generates
$O(ndk^2)$ predicates and $O(nd^2k^2)$ rules
where \(k\) is the fixed upper bound on the number of DNF clauses per local function,
the maximum being \(\binom d {\lfloor d/2\rfloor}\).
With this maximum value,
the number of solutions matches with the number of distinct monotonic Boolean functions,
the Dedekind number\cite{Kleitman1969}, currently known up to
\(d=8\)\cite{Wiedemann1991}\footnote{for $0\leq d\leq 8$: 2, 3, 6, 20, 168, 7581, 7828354, 2414682040998,
56130437228687557907788}.
Whenever the specified \(k\) is lower than the maximum, Boolean functions are not captured
by the encoding.
The constraints on canonicity are necessary to obtain efficient enumeration of solutions.
Whenever checking only for the existence of at least one solution, these constraints can be
relaxed, reducing the number of predicates and rules to \(O(ndk)\).

We detail the encoding hereby.
We use a predicate template
\lstinline|clause(N,C,L,S)| to specify that the literal \lstinline|L| with sign
\lstinline|S| is included in the \lstinline|C|-th clause of the DNF of \(f_{\text{\lstinline|N|}}\).
For instance, the two-clauses DNF $f_a(x)=(\lnot x_a \land x_b) \lor x_c$ is encoded by the three following predicates: \lstinline|clause(a,1,a,-1)|, \lstinline|clause(a,1,b,1)| and \lstinline|clause(a,2,c,1)|.

The domain of arguments \lstinline|N|, \lstinline|L|, and \lstinline|S| is fully determined by the
input influence graph \((V,E_+,E_-)\);
\lstinline|C| ranges from $1$ to $k$.
The influence graph is encoded with \lstinline|node/1| predicates with
\lstinline|node($i$)| if and only if \(i\in V\), and
\lstinline|in/3| predicates such that
\lstinline|in($j$,$i$,1)| if and only if $(j,i)\in E_+$ and
\lstinline|in($j$,$i$,-1)| if and only if $(j,i)\in E_-$.
The bound on the number of clauses is set by \lstinline|maxC(N,$k$)|:
\begin{lstlisting}
{clause(N,1..C,L,S): in(L,N,S), maxC(N,C)}.
\end{lstlisting}
The local monotonicity is ensured by denying a literal appearing with both
signs in the DNF of each component \lstinline|N|:
\begin{lstlisting}
:- clause(N,_,L,S), clause(N,_,L,-S).
\end{lstlisting}
DNFs without clauses result in constant functions, specified with the predicate
\lstinline|constant/2|:
\begin{lstlisting}
1 {constant(N,(-1;1))} 1 :- node(N),          not clause(N,_,_,_).
\end{lstlisting}

The canonicity is obtained by ensuring the clauses are ordered by size and then lexicographically,
and without subset relation.
The ordering by size is guaranteed by the following integrity constraints.
The first line ensures that clauses identifiers increase continuously from \(1\).
\begin{lstlisting}
:- clause(N,C,_,_), not clause(N,C-1,_,_),    C > 1.
size(N,C,X) :- clause(N,C,_,_),               X = #count{L,S: clause(N,C,L,S)}.
:- size(N,C1,X1), size(N,C2,X2), X1 < X2,     C1 > C2.
\end{lstlisting}
The lexicographic ordering between clauses of the same size is enforced as follows, where
\lstinline|clausediff(N,C1,C2,L)| indicates that \lstinline|L| is present in the \lstinline|C1|-th
clause but not in the \lstinline|C2|-th;
and \lstinline|mindiff(N,C1,C2,L)| indicates that \lstinline|L| is the smallest literal such that
\lstinline|clausediff(N,C1,C2,L)|.
\begin{lstlisting}
:- size(N,C1,X), size(N,C2,X), C1 > C2, mindiff(N,C1,C2,L1), mindiff(N,C2,C1,L2), L1 < L2.
clausediff(N,C1,C2,L) :- clause(N,C1,L,_),    not clause(N,C2,L,_), clause(N,C2,_,_).
mindiff(N,C1,C2,L) :- clausediff(N,C1,C2,L),  L <= L' : clausediff(N,C1,C2,L'); clause(N,C1,L',_).
\end{lstlisting}
Finally, the absence of subset relation is guaranteed by the following integrity constraint:
\begin{lstlisting}
:- size(N,C1,X1), size(N,C2,X2), X1 <= X2,    clause(N,C2,L,S): clause(N,C1,L,S);       C1 != C2.
\end{lstlisting}

\subsection{Evaluation of Boolean functions}

We define generic rules to evaluate Boolean functions on meta-configurations.
A meta-configuration is specified similarly to configurations, with predictates
\lstinline|mcfg(H,N,V)|, where \lstinline|V| in \(\{-1,1\}\), but with potentially
two predicates \lstinline|mcfg($h$,$i$,-1)| \lstinline|mcfg($h$,$i$,1)| indicating that
the component \(i\) is free in the meta-configuration \(h\), i.e., \(h_i=*\).
The encoding of dynamical constraints takes care about instantiating their related
\lstinline|mcfg/3|.

The rules ensure that \lstinline|eval($h$,$i$,1)| (resp.
\lstinline|eval$(h$,$i$,-1)|) if and only if there exists a configuration \(x\in c(h)\) such that
\(f_i(x)\) is true (resp. false).
A clause is evaluated to false whenever one of its literal evaluates to false
(\pref{line:eval-clause-neg}); and to true whevener all its literals evaluate to true
(\pref{line:eval-clause-pos}).
Then, either the function is a constant and its evaluation follows the constant value
(\pref{line:eval-constant}), or the function is evaluated to true if all its clauses have been
evaluated true (\pref{line:eval-pos}); and to false whenever one ot its clauses is evaluated false
(\pref{line:eval-neg}).

\begin{lstlisting}
eval(H,N,C,-1) :- clause(N,C,L,-V), mcfg(H,L,V).%\label{line:eval-clause-neg}
eval(H,N,C,1) :- clause(N,C,_,_), mcfg(H,_,_), mcfg(H,L,V): clause(N,C,L,V).%\label{line:eval-clause-pos}
eval(H,N,1) :- eval(H,N,C,1); clause(N,C,_,_).%\label{line:eval-pos}
eval(H,N,-1) :- clause(N,_,_,_), mcfg(H,_,_), eval(H,N,C,-1): clause(N,C,_,_).%\label{line:eval-neg}
eval(H,N,V) :- constant(N,V), mcfg(H,_,_).%\label{line:eval-constant}
\end{lstlisting}

For each meta-configuration, this encoding generates \(O(nk)\) predicates and \(O(ndk)\) rules.

\subsection{Positive Reachability}

Each \((m,m')\in\PR\) is translated as a predicate \lstinline|reach($m$,$m'$)|, specifying that the
configuration \(x^m\) has to be able to reach the configuration \(x^{m'}\).

Following \pref{def:reachability}, reachability properties in most permissive BNs can be assessed
with particular meta-configurations.
The rule below declares a meta-configuration dedicated to the positive reachability
constraint, initially being equal to the initial configuration.
\begin{lstlisting}
mcfg((pr,X,Y),N,V) :- reach(X,Y), cfg(X,N,V).
\end{lstlisting}
Then, the meta-configuration has to be extended to satisfy the (constrained) trap space property
(\pref{def:constrained-trap-space}).
The extensions of meta-configurations are encoded with \lstinline|ext(H,N,V)| predicates, and their
application is encoded by the generic rule in \pref{line:mcfg-ext}.
Whenever the function of the component \lstinline|N| of the meta-configuration can be evaluated to
its value in the target configuration \lstinline|Y|, the meta-configuration is extended to include
this value (\pref{line:pr_ext-fix}).
Whenever the function can be evaluated to the opposite value of the target configuration, its
inclusion in the meta-configuration is a choice (\pref{line:pr_ext-choice}).
\begin{lstlisting}
mcfg(H,N,V) :- ext(H,N,V).%\label{line:mcfg-ext}
ext((pr,X,Y),N,V) :- reach(X,Y), eval((pr,X,Y),N,V), cfg(Y,N,V).%\label{line:pr_ext-fix}
{ext((pr,X,Y),N,V)} :- reach(X,Y), eval((pr,X,Y),N,V), cfg(Y,N,-V).%\label{line:pr_ext-choice}
\end{lstlisting}
The resulting meta-configuration is a \(L\)-constrained trap space, where \(L\) is the set of
components where the extensions of \pref{line:pr_ext-choice} have been skipped, provided the
opposite value is not already in the initial configuration.

Finally, the two properties that the constrained trap space has to verify (\pref{def:reachability})
lead to the following rules.
The first rejects models where the target configuration is not included in the meta-configuration;
the second rejects models where a component is free in the meta-configuration (therefore not in
\(L\)), but its target value can not be obtained with its function in the scope of the constrained
trap space.
\begin{lstlisting}
:- cfg(Y,N,V), not mcfg((pr,X,Y),N,V), reach(X,Y).
:- cfg(Y,N,V), not ext((pr,X,Y),N,V), ext((pr,X,Y),N,-V), reach(X,Y).
\end{lstlisting}

Accounting for \lstinline|eval|-related rules, for each \lstinline|reach(X,Y)| predicate, \(O(nk)\) predicates and \(O(ndk)\) rules are generated.

\subsection{Negative Reachability}

Each \((m,m')\in\NR\) is translated as a predicate \lstinline|nonreach($m$,$m'$)|, specifying that
it is impossible to reach the configuration \(x^{m'}\) from configuration \(x^m\).

The most permissive reachability property recalled in \pref{def:reachability} relies on the
\emph{existence} of subset of components \(L\subseteq\range 1n\) so that the smallest \(L\)-constrained trap
space \(w\) containing the initial configuration (1) contains the target configurations, 
and (2) for each component \(i\) not in \(L\), there exists a configuration \(z\in c(w)\) such that
\(f_i(z)=y_i\).

Proving the absence of reachability would require that these conditions are verified by none of
these subsets of components \(L\).
In \cite{mpbn-tr}, it has been demonstrated that its is sufficient to consider at most \(n\)
particular subsets of components \(L\) to conclude on the absence of reachability.
Essentially, we start verifying the conditions with \(L=\emptyset\) and then iteratively add
in \(L\) the components which do not satisfy the condition (2).
With this procedure, it is sufficient to check the condition (1) in the \(L\) obtained at the
\(n^{\mathrm{th}}\) iteration.

To assess the non-reachability of configuration \(y\) from \(x\), our encoding generates
\(n\) meta-configurations, initially being equal to \(x\) (\pref{line:iter}-\ref{line:mcfg-nr}).
Then predicates \lstinline|locked(X,Y,I+1,N)| specify that the component \lstinline|N| is in the
\lstinline|I+1|$^\mathrm{th}$ iteration of \(L\).
Such a predicate has to be true if \lstinline|N| does not verify condition (2) at iteration
\lstinline|I| (\pref{line:nr-lock}), or if it is already in \(L\) at the preceding iteration
(\pref{line:lock-cont}).
The extension of the meta-configuration at iteration \lstinline|I| is then constrained by components
in \(L\) (\pref{line:nr-ext}).
Finally, if there exists a component \lstinline|N| such that \(y_{\text{\lstinline|N|}}\) is not the
meta-configuration of the last iteration, the predicate \lstinline|nr($x$,$y$)| is true, indicating the
absence of reachability (\pref{line:nr-ok}).
A model is rejected if such a predicate cannot be proven true (\pref{line:nr}).

\begin{lstlisting}
iter(1..K) :- nbnode(K).%\label{line:iter}
mcfg((nr,X,Y,I),N,V) :- nonreach(X,Y), cfg(X,N,V), iter(I).%\label{line:mcfg-nr}
locked(X,Y,I+1,N) :- cfg(X,N,V), cfg(Y,N,V), not ext((nr,X,Y,I),N,V), ext((nr,X,Y,I),N,-V), iter(I+1).%\label{line:nr-lock}
locked(X,Y,I+1,N) :- locked(X,Y,I,N), iter(I+1).%\label{line:lock-cont}
ext((nr,X,Y,I),N,V) :- not locked(X,Y,I,N), eval((nr,X,Y,I),N,V).%\label{line:nr-ext}
nr(X,Y) :- not mcfg((nr,X,Y,K),N,V), nbnode(K), cfg(Y,N,V), nonreach(X,Y).%\label{line:nr-ok}
:- not nr(X,Y), nonreach(X,Y).%\label{line:nr}
\end{lstlisting}

Accounting for \lstinline|eval|-related rules, for each \lstinline|nonreach(X,Y)| predicate, this
encoding generates \(O(n^2k)\) predicates and \(O(n^2dk)\) rules.

\subsection{Attractors}

As indicated in \pref{sec:problem}, we consider two different properties related to the attractors
of the BN \(f\):
fixpoints properties, where specified configurations have to be fixpoints of \(f\);
and trap space properties, where specified configurations have to belong to trap spaces where a subset
of their components have a fixed value.

Accounting for \lstinline|eval|-related rules,
the encoding of each of the following properties
generates \(O(nk)\) predicates and \(O(ndk)\) rules.

\subsubsection{Fixpoints}

Each \(m\in\FP\) is translated as a predicate \lstinline|is_fp($m$)|, specifying that
the configuration \(x^{m}\) is a fixpoint of \(f\) .
The constraint is ensured by rejecting models where the evaluation gives an
opposite value for at least one component:
\begin{lstlisting}
mcfg(X,N,V) :- is_fp(X), cfg(X,N,V).
:- is_fp(X), cfg(X,N,V), eval(X,N,-V).
\end{lstlisting}

\subsubsection{Trap spaces}

Each \((m,i)\in\TP\) is translated as a predicate \lstinline|is_tp($m$,$i$)|, specifying that
the smallest trap space \(t\) containing the configuration \(x^m\)
has to have the component \(i\) fixed, i.e., \(t_i\neq *\).
The initialisation and extension of the smallest trap space containing \(x\) are obtained with rules
in \pref{line:tp-init}-\ref{line:tp-ext}.
The model is rejected if the resulting trap space has any free component specified as trapped.
\begin{lstlisting}
mcfg((ts,X),N,V) :- cfg(X,N,V), is_tp(X,_).%\label{line:tp-init}
mcfg((ts,X),N,V) :- eval((ts,X),N,V).%\label{line:tp-ext}
:- is_tp(X,N), cfg(X,N,V), mcfg((ts,X),N,-V).
\end{lstlisting}

\section{Evaluation}
\label{sec:evaluation}

We performed experiments to assess the scalability and illustrate potential biological applications of our encoding of BN synthesis.
We used the ASP solver \textsc{Clingo}\footnote{version 5.3.0 available at
\url{https://potassco.org/clingo}} using default solving strategies\footnote{Instances
available at \url{http://www.labri.fr/perso/lpauleve/ictai19.zip}}.

\subsection{Scalability on Random Boolean Networks}
\label{sec:experiments}
We randomly generated scale-free directed graphs with different biases on the in-degree of nodes in
order to obtain influence graphs similar to the usually encountered with gene and cell signalling
networks.

\begin{figure}[!t]
    \centering
    \includegraphics[width=0.8\linewidth]{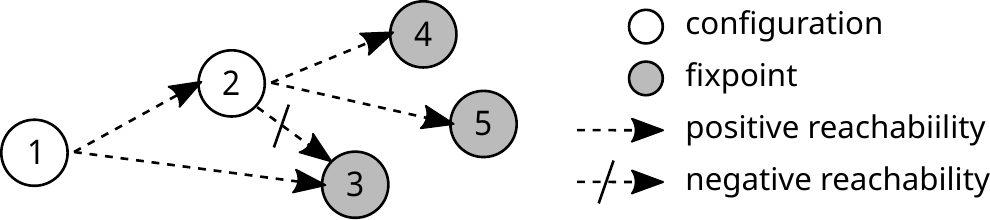}
    \caption{Sketch of the constraints for the synthesis on random
    graphs}
    \label{fig:sketch-random-prop}
\end{figure}

The synthesis has then been performed with each of these networks as input influence graph, and
with a generic dynamical property of a two stages differentiation processes, as illustrated
in \pref{fig:sketch-random-prop}.
The properties are specified using 5 \emph{empty} observations \(\range15\), among which 3 should
match with a distinct fixpoint (\(\FP=\{3,4,5\}\)).
The first observation is supposed to reach the second and third, whereas the second is expected to
reach the fourth and fifth, but not the third:
\(\PR=\{(1,2),(1,3),(2,4),(2,5)\}\), \(\NR=\{(2,3)\}\).

\begin{figure}[!t]
    \includegraphics[width=\linewidth]{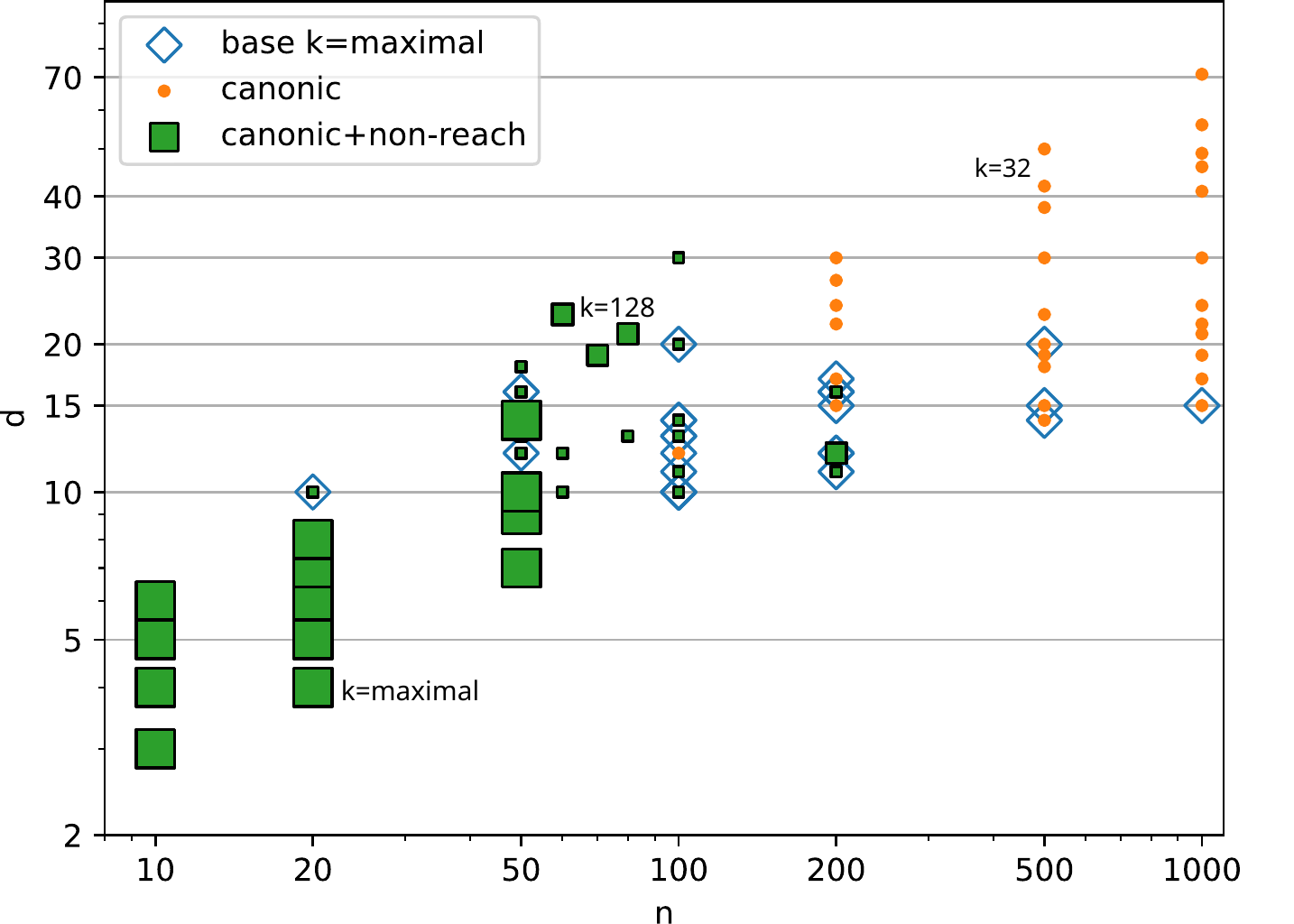}
    \caption{Successfully solved random instances for different sets of constraints (marker shapes)
        and different bounds on the number of clauses (marker sizes) in function of the number of components \(n\) and maximal
in-degree \(d\).}
    \label{fig:exp-random-solve}
\end{figure}

\pref{fig:exp-random-solve} gives an overview of successfully solved instances within 2h of CPU time
(2.5Ghz).
With canonic solutions and the maximal number of clauses, it scales to networks up to 50
nodes, with maximal in-degree \(15\).
With bounded number of clauses, instances with up to \(200\) nodes have been solved, provided a
similar in-degree.
Solving larger instances requires dropping negative reachability constraints.
The main limit is the number of variables and rules generated by the encoding, which is often larger
than \(2^{32}\) with negative reachability.
Almost all solved instances are satisfiable, except in a couple of cases with \(\leq 20\) nodes with
negative reachability.

\subsection{Application to Cell Differentiation Modelling}
\label{sec:application}

\begin{figure}[!t]
\begin{center}
\includegraphics[width=7cm]{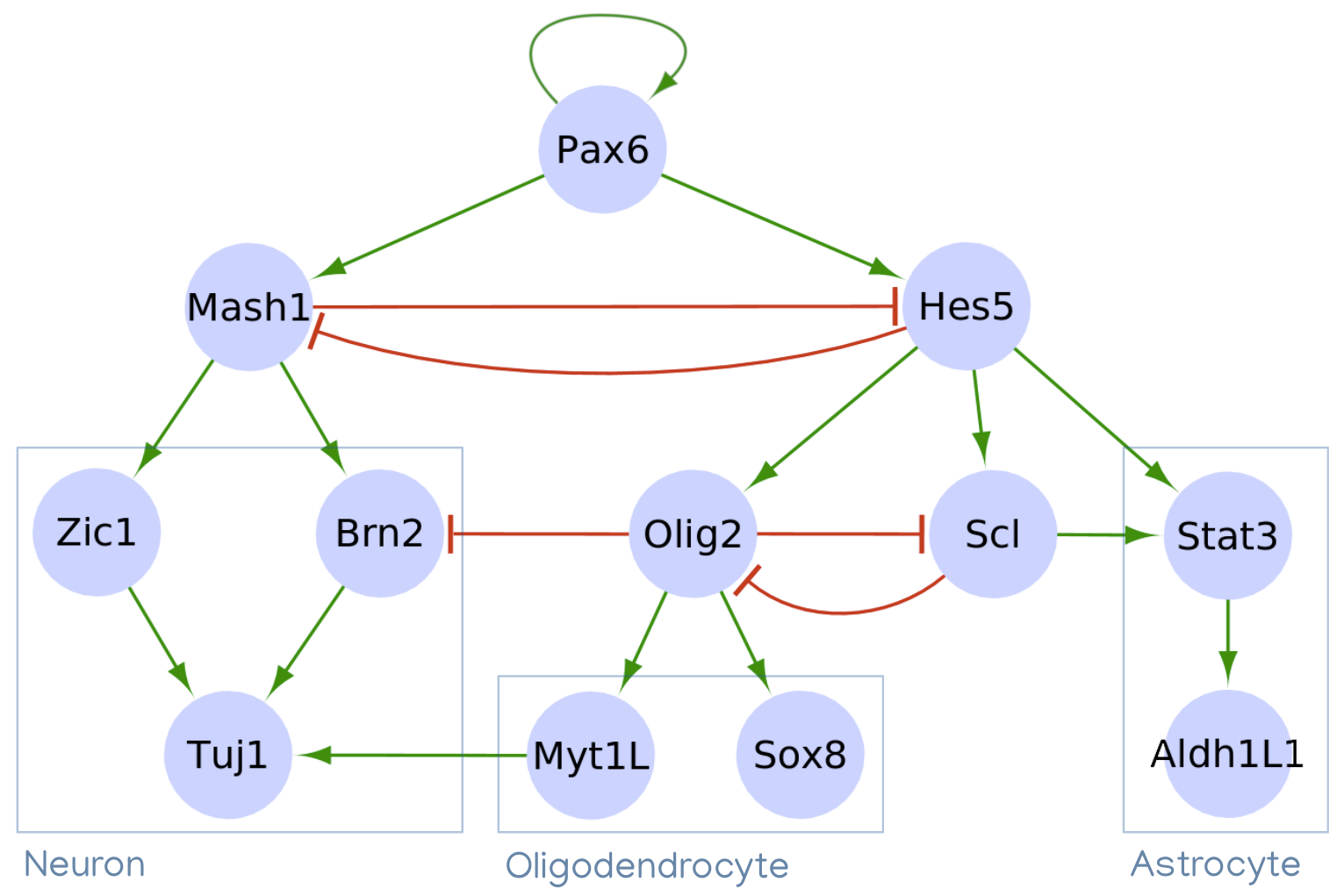}
\caption{Influence graph for CNS development}
\label{fig:PKN_Qiu}
\end{center}
\end{figure}

We illustrate our methodology on a cell differentiation context: the central nervous system (CNS) development.
Neural stem cells can terminally differentiate into neurons, astrocytes and
oligodendrocytes, and an influence graph gathering known gene interactions is available in the
literature\cite{Qiu_2017}. This graph with two differentiation stages (Fig.~\ref{fig:PKN_Qiu}) consists of 12 genes.
Despite its relatively small size, this influence graph already entails more than 226 millions of compatible BNs
(the number of BNs compatible with an influence graph is given by the product of the Dedekind numbers related to each node).

\begin{table}[!t]
\caption{List of observed nodes in each observation}
\label{tab:obsQiu}
\begin{tabular}{|M{1cm}|m{2.5cm}|m{4cm}|}
\hline
\textbf{obs. ID} & \textbf{activated genes} & \textbf{inactivated genes}\\\hline
\emph{0}		& \textit{none} & \textit{all} \\
\emph{iPax6}	& Pax6			& \textit{the 11 others} \\
\emph{tM}		& Pax6			& Aldh1L1, Olig2, Scl, Sox8, Tuj1 \\
\emph{fT}		& Brn2, Tuj1, Zic1	& Aldh1L1, Sox8 \\
\emph{tO}		& Olig2, Pax6	& Aldh1L1, Scl, Sox8, Tuj1 \\
\emph{fMS}		& Sox8			& Aldh1L1, Brn2, Tuj1, Zic1 \\
\emph{tS}		& Pax6, Scl		& Aldh1L1, Olig2, Sox8, Tuj1 \\
\emph{fA}		& Aldh1L1		& Brn2, Sox8, Tuj1, Zic1 \\\hline
\end{tabular}
\end{table}
The observations are given in Table~\ref{tab:obsQiu}, and the positive and negative reachability
constraints are set as
\(\PR = \{
    (\text{\it iPax6}, \text{\it tM}),\) 
    \( (\text{\it tM}, \text{\it fT}), \)
    \(    (\text{\it iPax6}, \text{\it tO}),\) \((\text{\it tO}, \text{\it fMS}),
    (\text{\it iPax6}, \text{\it tS}), (\text{\it tS}, \text{\it fA})\}
\),
\(\NR = \{
    (\text{\it 0}, \text{\it fT}), \) \(
        (\text{\it 0}, \text{\it fMS}),
        (\text{\it 0}, \text{\it fA})
    \}\).

To test the impact of various hypotheses on the stability of the phenotypes, trap spaces (with the
fixation of the 4 phenotypes markers Aldh1L1, Myt1L, Sox8 and Tuj1) and fixpoints constraints are applied on the observations \emph{fT}, \emph{fMS} and \emph{fA}.
Their relevance depends on the assumptions and knowledge precision about the phenotypes.

To appreciate the pertinence of the method, Table~\ref{tab:bnQiu} presents the number of inferred BNs given each defined constraint and combinations thereof.
\begin{table}[!t]
\begin{center}
\caption{Number of admissible BNs w.r.t. various properties}
\label{tab:bnQiu}
\begin{tabular}{|l|p{2.3cm}|}
\hline
\textbf{applied constraints} & \textbf{\# solutions}\\\hline
\multicolumn{2}{|c|}{application of a single type of constraint:}\\\hline
3 negative reachability \emph{(NR)}						&	224\,025\,280	\\
6 positive reachability \emph{(PR)}							&	24\,076\,416 	\\
12 trap spaces	\emph{(TP)}							&	17\,220\\
3 fixpoints \emph{(FP)}								&	4970\\\hline
\multicolumn{2}{|c|}{application of combination of constraints:}\\\hline
\emph{PR} + \emph{NR}							&	16\,050\,944	\\
\emph{PR} + \emph{TP}							&	8964	\\
\emph{NR} + \emph{TP}							&	5667	\\
\emph{PR} + \emph{NR} + \emph{TP}   			&	3735\\
\emph{PR} + \emph{FP} 							&	3360\\
\emph{PR} + \emph{NR} + \emph{FP}	&	1120	\\\hline
\end{tabular}
\end{center}
\end{table}
Each constraint complements the filtering by adding new information, and
while 226 millions of BNs were candidates for modelling the CNS development, applying a relevant combination of constraints leads to select almost instantaneously the relatively small set of models respecting the observed behaviours. This huge reduction combined with the exhaustiveness of the method is twice interesting for biological studies. It first enables the analysis of variability across the models to study the significance of the components in the observed behaviours. Secondly, it offers the opportunity to quantify the data informativeness and even inform of the inconsistency of an hypothesis.

\section{Discussion}
\label{sec:conclusion}

Taking advantage of stable models offered by ASP, we provide a compact encoding of the BN synthesis
from static and dynamical properties, with part of the complexity being parametrized.
The method enables addressing scales and type of dynamical properties beyond the scope of already
existing approaches.

Although not explicitly addressed in the encoding and evaluation, the use of ASP also enables
efficient synthesis with optimization, e.g., finding BNs with minimal/maximal influence graph.

Negative reachability has a limited scalability due to the \(O(n^2)\) variables and rules it
generates (\(n\) being the dimension of the BNs).
Future work will investigate SMT-like approaches to generate part of the constraints on the fly.

The considered properties are inspired by models of cellular differentiation. In such a context,
having access to the complete set of candidate models enables uncovering influence motifs which are
key for reproducing desired behaviours.
Related to the applications, being able to account for universal properties on (reachable)
attractors in the synthesis would increase the precision of inferred models, and constitutes a
challenging direction.

\section*{Acknowledgement}
Part of the experiments was
carried out using the PlaFRIM experimental
testbed, supported by Inria, CNRS (LABRI and IMB), Universit\'e de Bordeaux, Bordeaux INP and Conseil
R\'egional d'Aquitaine (see \url{https://www.plafrim.fr}).

\bibliographystyle{IEEEtran}
\bibliography{bn-synthesis}

\begin{thebibliography}{10}
\providecommand{\url}[1]{#1}
\csname url@samestyle\endcsname
\providecommand{\newblock}{\relax}
\providecommand{\bibinfo}[2]{#2}
\providecommand{\BIBentrySTDinterwordspacing}{\spaceskip=0pt\relax}
\providecommand{\BIBentryALTinterwordstretchfactor}{4}
\providecommand{\BIBentryALTinterwordspacing}{\spaceskip=\fontdimen2\font plus
\BIBentryALTinterwordstretchfactor\fontdimen3\font minus
  \fontdimen4\font\relax}
\providecommand{\BIBforeignlanguage}[2]{{%
\expandafter\ifx\csname l@#1\endcsname\relax
\typeout{** WARNING: IEEEtran.bst: No hyphenation pattern has been}%
\typeout{** loaded for the language `#1'. Using the pattern for}%
\typeout{** the default language instead.}%
\else
\language=\csname l@#1\endcsname
\fi
#2}}
\providecommand{\BIBdecl}{\relax}
\BIBdecl

\bibitem{concurrency-in-BNs}
T.~Chatain, S.~Haar, J.~Kol{\v c}{\'a}k, L.~Paulev{\'e}, and A.~Thakkar,
  ``{Concurrency in Boolean networks},'' \emph{{Natural Computing}}, 2019.

\bibitem{Terfve2012}
C.~Terfve, T.~Cokelaer, D.~Henriques, A.~MacNamara, E.~Goncalves, M.~K. Morris,
  M.~v. Iersel, D.~A. Lauffenburger, and J.~Saez-Rodriguez, ``{CellNOptR}: a
  flexible toolkit to train protein signaling networks to data using multiple
  logic formalisms,'' \emph{BMC Systems Biology}, vol.~6, no.~1, p. 133, 2012.

\bibitem{Dorier2016}
J.~Dorier, I.~Crespo, A.~Niknejad, R.~Liechti, M.~Ebeling, and I.~Xenarios,
  ``Boolean regulatory network reconstruction using literature based knowledge
  with a genetic algorithm optimization method,'' \emph{BMC Bioinformatics},
  vol.~17, no.~1, p. 410, 2016.

\bibitem{Caspots-BioSystems16}
M.~Ostrowski, L.~Paulev{\'e}, T.~Schaub, A.~Siegel, and C.~Guziolowski,
  ``Boolean network identification from perturbation time series data combining
  dynamics abstraction and logic programming,'' \emph{Biosystems}, vol. 149,
  pp. 139 -- 153, 2016.

\bibitem{yordanov2016a}
B.~Yordanov, S.-J. Dunn, H.~Kugler, A.~Smith, G.~Martello, and S.~Emmott, ``A
  method to identify and analyze biological programs through automated
  reasoning,'' \emph{Systems Biology and Applications}, vol.~2, 2016.

\bibitem{Kauffman69}
S.~A. Kauffman, ``Metabolic stability and epigenesis in randomly connected
  nets,'' \emph{Journal of Theoretical Biology}, vol.~22, pp. 437--467, 1969.

\bibitem{Thomas73}
R.~Thomas, ``Boolean formalization of genetic control circuits,'' \emph{Journal
  of Theoretical Biology}, vol.~42, no.~3, pp. 563 -- 585, 1973.

\bibitem{Aracena09}
J.~Aracena, E.~Goles, A.~Moreira, and L.~Salinas, ``On the robustness of update
  schedules in {B}oolean networks,'' \emph{Biosystems}, vol.~97, no.~1, pp. 1
  -- 8, 2009.

\bibitem{beyond-general}
T.~Chatain, S.~Haar, and L.~Paulev{\'e}, ``{Boolean Networks: Beyond
  Generalized Asynchronicity},'' in \emph{{Cellular Automata and Discrete
  Complex Systems}}, ser. LNCS, vol. 10875.\hskip 1em plus 0.5em minus
  0.4em\relax {Springer}, 2018, pp. 29--42.

\bibitem{mpbn-tr}
T.~{Chatain}, S.~{Haar}, and L.~{Paulev{\'e}}, ``{Most Permissive Semantics of
  Boolean Networks},'' \emph{CoRR}, vol. abs/1808.10240, 2018.

\bibitem{baral02a}
C.~Baral, \emph{Knowledge Representation, Reasoning and Declarative Problem
  Solving}.\hskip 1em plus 0.5em minus 0.4em\relax Cambridge University Press,
  2003.

\bibitem{gekakasc12a}
M.~Gebser, R.~Kaminski, B.~Kaufmann, and T.~Schaub, \emph{Answer Set Solving in
  Practice}, ser. Synthesis Lectures on Artificial Intelligence and Machine
  Learning.\hskip 1em plus 0.5em minus 0.4em\relax Morgan and Claypool
  Publishers, 2012.

\bibitem{Lin2004}
F.~Lin and Y.~Zhao, ``{ASSAT}: Computing answer sets of a logic program by
  {SAT} solvers,'' \emph{Artificial Intelligence}, vol. 157, no.~1, pp.
  115--137, 2004.

\bibitem{Kleitman1969}
D.~Kleitman, ``On {D}edekind{\textquotesingle}s problem: The number of monotone
  {B}oolean functions,'' \emph{Proceedings of the American Mathematical
  Society}, vol.~21, no.~3, p. 677, 1969.

\bibitem{Wiedemann1991}
D.~Wiedemann, ``A computation of the eighth dedekind number,'' \emph{Order},
  vol.~8, no.~1, pp. 5--6, 1991.

\bibitem{Qiu_2017}
X.~Qiu, Q.~Mao, Y.~Tang, L.~Wang, R.~Chawla, H.~A. Pliner, and C.~Trapnell,
  ``Reversed graph embedding resolves complex single-cell trajectories,''
  \emph{Nature Methods}, vol.~14, no.~10, pp. 979--982, 2017.

\end{thebibliography}

\end{document}